\documentclass[a4paper,twocolumn,10pt,times]{article}
 \usepackage{epsfig}
\begin{document}

\date{}    

\title{\bf Fusion for Evaluation of Image Classification in Uncertain Environments}


\author{
A.  Martin \\
$\mbox{E}^3\mbox{I}^2$ EA3876 \\
ENSIETA \\
2 rue Fran{\c c}ois Verny, 29806 Brest Cedex 09, France \\
Arnaud.Martin@ensieta.fr\\}

\maketitle                        
\thispagestyle{empty}


\noindent
{\bf Abstract -
   {\small\em We present in this article a new evaluation method for classification and segmentation of textured images in uncertain environments. In uncertain environments, real classes and boundaries are known with only a partial certainty given by the experts. Most of the time, in many presented papers, only classification or only segmentation are considered and evaluated. Here, we propose to take into account both the classification and segmentation results according to the certainty given by the experts. We present the results of this method on a fusion of classifiers of sonar images for a seabed characterization.}
}

\vspace{0.5cm}


\section{Introduction}

Textured image classification is a difficult problem in image processing and it is fundamental for a lot of applications. Many features can be extracted from the images to classify, and many classification algorithms can be used \cite{Russ02}. Hence, it is really necessary to evaluate their performance in order to compare them and choose the most adapted to the application. 

For instance, with satellite or sonar images, human experts must be able to classify the types of soils present in the images. Many types of soils can be encountered in a single image, and classification must be done on a local part of the image (pixel-wise, or often on small tiles of {\em e.g.} 16 $\times$ 16 or 32 $\times$ 32 pixels) taken as unit for the classification algorithm. Hence, after the image classification, an implicit image segmentation is obtained according to the size of the tiles. One image will be segmented into several patches, each one corresponding to a class ({\em e.g.} a specific type of soil).

The image classification methods are currently evaluated by the confusion matrix. Good-classification rates and error rates are usually calculated from this matrix. We must know the real class of the considered units of the images in order to establish the confusion matrix. Confusion matrix does not give an evaluation of the produced segmentation.

In order to evaluate the segmentation, we can not only consider visual comparison between the initial image and the segmented image. The image segmentation evaluation is still a studied problem \cite{Zhang96,Zhang97,Roman01,Mena05}. We can consider two cases: we do not have any {\it a priori} knowledge of the correct segmentation, or we have an {\it a priori} knowledge of the correct segmentation. Here we are in the second case because of the confusion matrix for which we need to get referenced images. In order to obtain these referenced images, experts must manually provide the image segmentation, for example {\it via} a visual inspection. Zhang in \cite{Zhang96} gives a review of usual discrepancy measures based on different distances between the segmented-pixel and the referenced-pixel. Most of the time, only one measure of mis-segmented pixel is given. We will propose on the contrary, in this article a linked study of one well-segmented pixel measure and a mis-segmented pixel measure. Indeed, in general case, if a pixel is not mis-segmented, it is not necessary well-segmented. So we can have few mis-segmented pixels but also few well-segmented pixels: the segmentation is not good.

We think that global image classification evaluation must be made by evaluating both the classification on considered units (with the confusion matrix) and in the same time by the evaluation of the produced segmentation (well-segmented pixel measure and a mis-segmented pixel measure) \cite{Martin06}. 

\begin{figure}[htb]
\vspace{-0.5cm}
\begin{center}
\includegraphics[height=5cm]{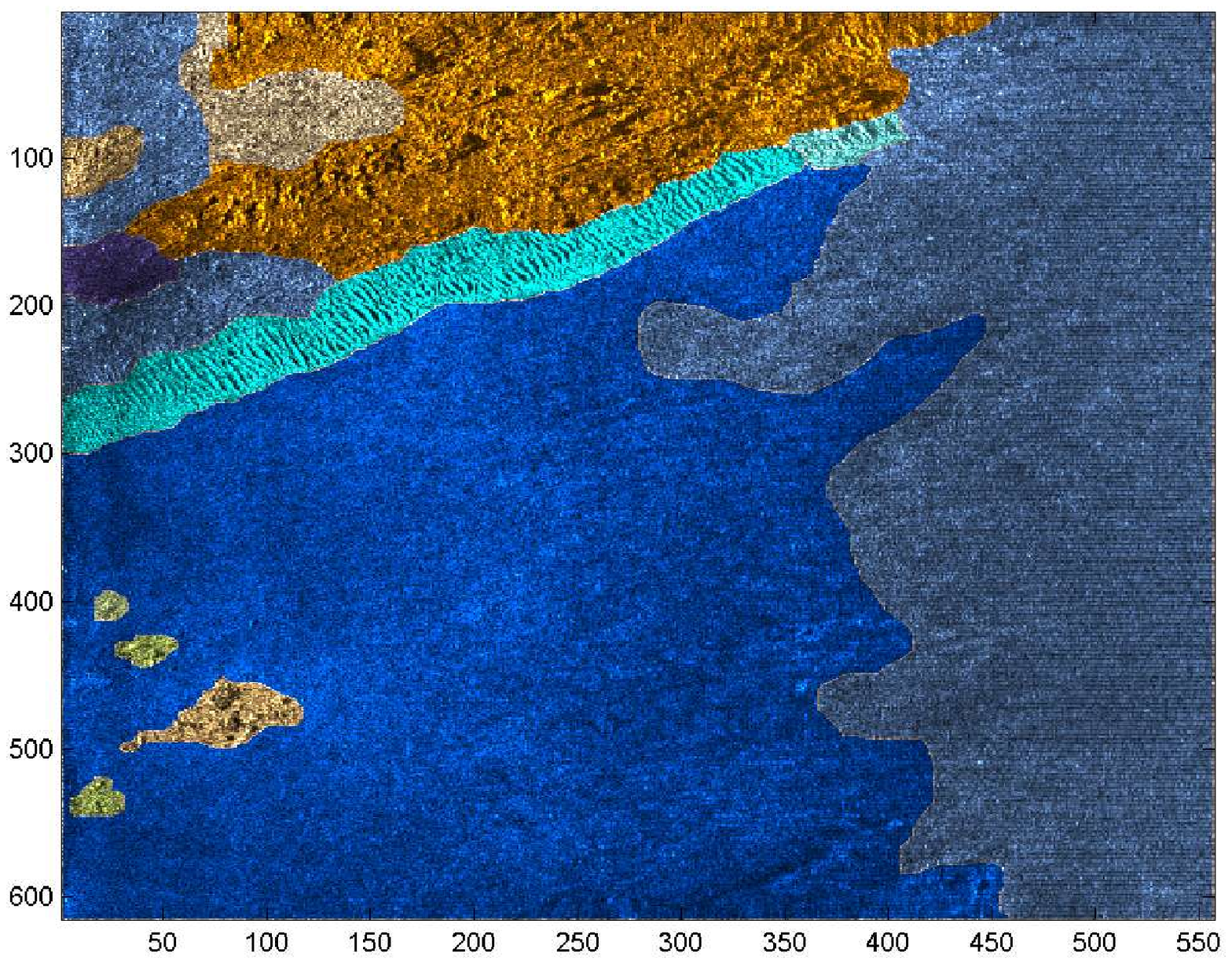}
\includegraphics[height=5cm]{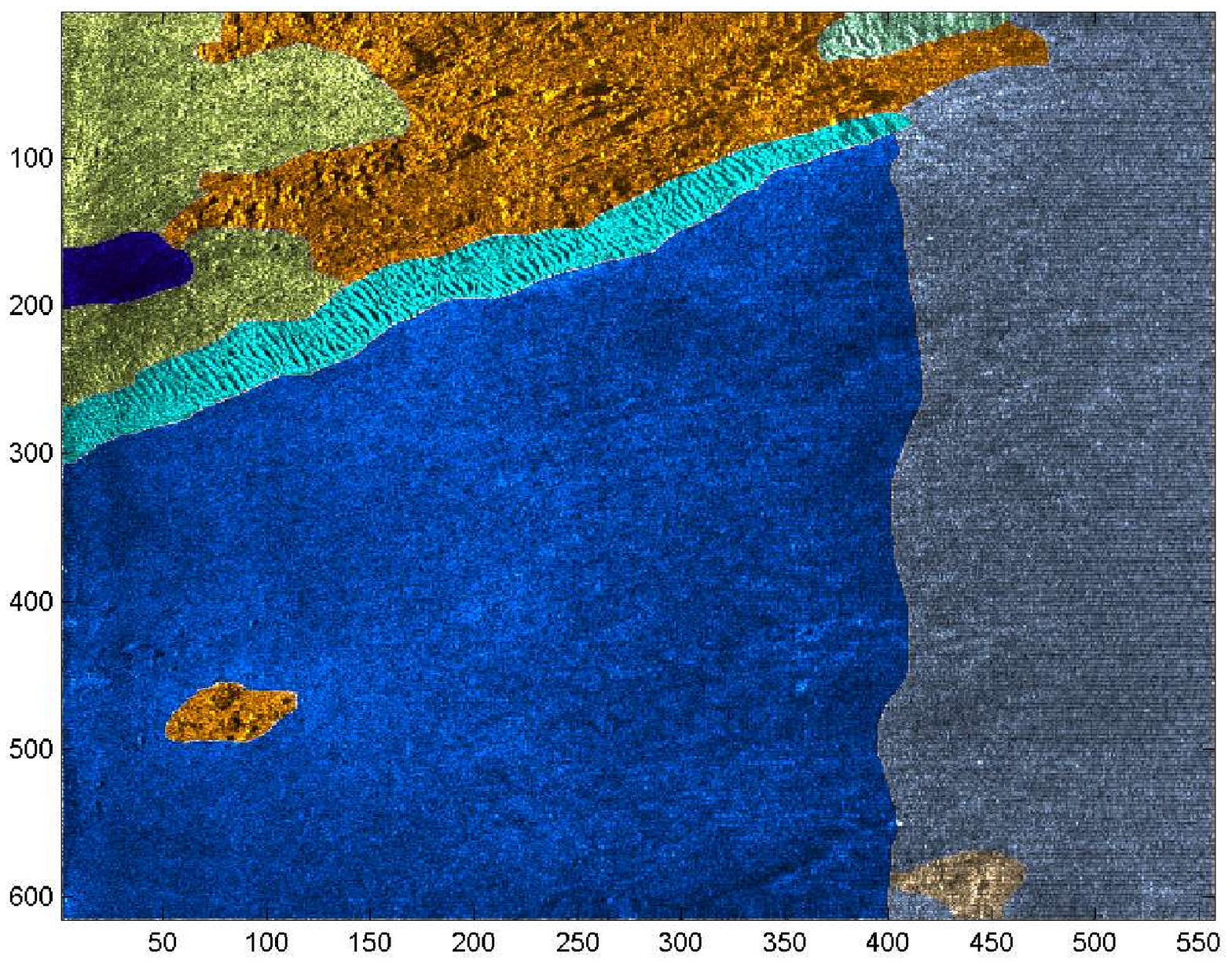}
\end{center}
\vspace{-0.5cm}
\caption{Segmentation given by two experts.}
\vspace{-0.5cm}
\label{expert}
\end{figure}

In real applications, it is really hard for one human expert to provide a certain information on the class and on the boundaries between the classes. For instance, the seabed characterization with sonar images cannot be made by human expert with a sufficient certainty. These images, illustrating this paper, are obtained with many imperfections \cite{Martin05}. Figure \ref{expert} exhibits the differences between the interpretation and the certainty of two sonar experts trying to differentiate the type of sediment (rock, cobbles, sand, ripple, silt) or shadow when the information is invisible (each color correspond to a kind of sediment and the associated certainty of the expert for this sediment expressed in terms of sure, moderately sure and not sure). 

We propose here a new approach for textured image classification and segmentation taking into account the information given by multiple experts and their certainty. In section \ref{classification}, we show how to integrate the expert certainty in confusion matrix and then deduce a good-classification rate and error classification rate, and how to fuse the different expert opinions. In section \ref{segmentation}, we propose two new distance-based measures in order to evaluate well and mis-segmented pixel taking into account the experts certainties. This evaluation is illustrated in section \ref{illustration} on real sonar images, in order to evaluate a fusion of the classifiers presented in \cite{Martin05}.

\section{Image classification evaluation}
\label{classification}

In this section, we propose an original evaluation approach for classification based on a new confusion matrix taking into account the uncertainty and the possibility that one unit belongs more than one class. This evaluation approach is adapted to the image classification evaluation, but can be used for any classifier evaluation.

\subsection{Classical Evaluation}

A first step of the classical classification evaluation can be made by comparing the results of the classifier to the reality. But in order to evaluate a classification algorithm, many different configurations and tests must be considered. Classification algorithms can yield many variable results depending on the sample. Most of the time, classification algorithms evaluation is conducted by the confusion matrix. 

Confusion matrix is composed by the number $cm_{ij}$ of elements of the class $i$ classified in the class $j$. In order to obtain rates making it easier to compare different size of databases, we normalize this confusion matrix by:
\begin{equation}
\label{NCM}
Ncm_{ij} = \frac{cm_{ij}}{\displaystyle \sum_{j=1}^N cm_{ij}}=\frac{cm_{ij}}{N_i}, 
\end{equation}
with $N$ the number of considered classes and $N_i$ the number of element from the true class $i$. From this normalized confusion matrix a good-classification rate vector can be written as:
\begin{equation}
\label{GCR}
GCR_i = Ncm_{ii}, 
\end{equation}
and an error classification rate vector as:
\begin{equation}
\label{ECR}
ECR_i = \frac{1}{2} \left(\sum_{j=1, j\neq i}^{N} Ncm_{ij}+  \sum_{i=1, i \neq j}^{N} \frac{Ncm_{ij}}{N-1} \right). 
\end{equation}

This error classification rate is the mean of the two errors corresponding to the elements from a given class $i$ classified in another class (first term), and corresponding to the elements classified in a given class $j$ being from another class $i$ (second term). We do not have to normalize the first term because of the normalization of the confusion matrix on the rows, but the second term must be normalized by the number of rows minus one (because of the $Ncm_{ii}$ term corresponds to the good-classification).

Thus image classification algorithms evaluation must be made not only on one image but on the whole images database. As a consequence, we have to consider a non-normalized confusion matrix on each image and normalize the sum of the matrix confusion on all images of the database.

\subsection{Evaluation with certainty given by each expert}

We consider here a general case where information is given by the expert on each pixel and the classification algorithm is made on an unit of $n \times n$ pixels. Hence on each unit, more than one class can be present. Generally, the classification algorithms can find only one of these classes. In order to take into account the inhomogeneous units, consider that if the classification algorithm finds one of these classes on the unit, the algorithm is right in the proportion of this found class in the $n \times n$ pixels-unit and it is wrong in the proportion of the other classes in the considered unit. For instance, imagine the case where the expert considers a tile of size $16 \times 16$ pixels and declares that on a part of the unit, 50 given pixels belong to class 1, and 206 other pixels belong to class 3. If the classification algorithm finds the unit belongs to class 1, the confusion matrix will be computed by the recurrence relations: $cm_{11} \gets cm_{11}+50/256$ and $cm_{31} \gets cm_{31}+206/256$. Hence the confusion matrix is not composed of integer numbers and $N_i$ is also not integer; but the sums of column are still integers.

If the expert can give the class with a certainty grade, we must not take equally two different grades in our classification evaluation. For instance, in sonar application, the operator can be sure that one part of the image as belonging to rock, and be totally doubtful on another part of the image. Classical confusion matrices suppose that the reality is perfectly known and that is rarely the case especially in image classification. We propose to graduate this difference of information by different weights corresponding to the different grades of certainty that are considered. In the confusion matrix, such weights could be integrated easily in the general sum. For example, consider three grades of certainty (sure, moderately sure and not sure), we can choose respectively the weights: 2/3, 1/2 and 1/3. If one expert labels a unit as belonging to the class 1 ({\em e.g.} rock), with a moderate certainty, and if the classification algorithm finds the class 1, considering the previous given weights, the confusion matrix will be updated such as:  $cm_{11} \gets cm_{11}+1/2$. If the classification algorithm finds the class 2 ({\em e.g.} sand) on the considered unit, the confusion matrix becomes $cm_{12} \gets cm_{12}+1/2$. Hence the sums of columns are not integer anymore.

In order to fuse the referenced images provided by different experts, we can compare the classified image with all the referenced images by the experts. Hence we obtain as many non-normalized confusion matrices as experts, and we can simply combine them by addition. This can be done also if the experts do not provide certainty, in such a case the weight is 1 for all units.

By the simple addition of the non-normalized confusion matrices, we weight the obtained results by the image size or the considered unit number.

In order to obtained rates, we normalize the obtained confusion matrix with equation (\ref{NCM}) and calculate the good-classification rate vector with equation (\ref{GCR}) and the error classification rate vector with equation (\ref{ECR}). Of course these rates are not percentages anymore. For instance, the good-classification rate is no longer the percentage of well classified units, because the weights given by the inhomogeneous units or by the expert certainty are rational. These newly obtained confusion matrix, good-classification rate and error classification rate give a good evaluation of classification taking into account the inhomogeneous units and certainty of the experts. This approach can be applied in every domain where we try to classify uncertain elements, and not only in image classification. 

\section{Segmentation Evaluation}

\label{segmentation}

Image classification provides an implicit image segmentation, the boundaries are given by the difference of classes between two adjacent tiles. A good image classification evaluation has to study this obtained image segmentation. 

Many approaches can be considered in order to obtain boundaries. This is not the subject of this paper and the following segmentation evaluation can be applied to all image segmentations given by boundaries as a succession of pixels.

We propose here a linked study of one well-segmented pixel measure and a mis-segmented pixel measure. Generally one of these measures is considered in the case with an {\it a priori} knowledge \cite{Zhang96,Kanungo95,Peli82}. 
The well-segmented pixel measure is a well-detection boundary measure and the mis-segmented pixel measure is a false detection boundary measure. We show how these two measures can take into account the uncertainty of the expert on the position and existence of the boundaries, assuming that each certainty grade is represented by a weight. 

\subsection{Well-detection boundary measure}

\begin{figure}[htb]
    \begin{center}
\includegraphics[width=7cm]{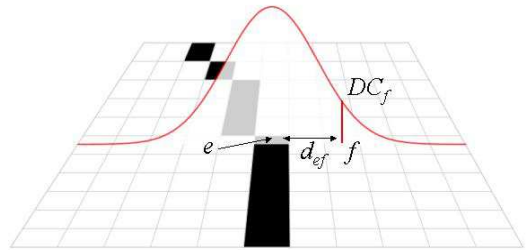}
    \end{center}
\caption{Distance weight for the well-detection criterion.}
\label{fig_pond}
\end{figure}

First, for each found boundary pixel $f$, search the minimal distance $d_{fe}$ between $f$ and all the boundary pixels provided by the expert $e$. Hence the pixel $e$ is a function of $f$, and we should note it as $e_f$, but in order to simplify notations, it is referred to as $e$ in the rest of the paper. We take here an Euclidean distance but any other distance can be envisaged. The certainty weight of the pixel $e$ given by the expert is noted as $W_e$. We define a well-detection criterion vector by:
\begin{equation}
\label{DC}
DC_f = \exp(-(d_{fe}.W_e)^2).W_e. 
\end{equation}
This criterion gives a Gaussian-kind distribution of weights with a standard deviation given by the certainty weights, as shown in figure \ref{fig_pond}.

The well-detection boundary measure is defined by the normalized well-detection criterion given by:
\begin{equation}
\label{WDC}
WDC = \frac{\sum_f DC_f} {\left(\max_f(DC_f) . \sum_e W_e\right)^a}.
\end{equation}
Hence, this measure is defined between 0 and 1. In real applications, this criterion remains small even for very good boundary detection, so we can take $a=1/6$ in order to accentuate small values. 

This criterion only takes into account {\it the distance} from the found boundary to the contour provided by the expert. However, the reference boundary has a local {\it direction} which is another aspect we have to consider. Indeed, for instance, a found boundary can cross a given boundary orthogonally: in this case some pixels from the found boundary are very near (in terms of distance) to pixels from the reference boundary but that is not a good detection. 

\begin{figure}[htb]
\begin{center}
\includegraphics[width=4cm]{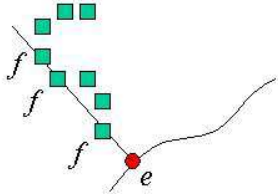}
\end{center}
\caption{Example of $n_{ef}$ for three given $f$, the found boundary is represented by green squares and the referenced boundary by a black line.}
\label{fig_n_ef}
\end{figure}

In order to take into account the local direction, we count, for a given pixel $f$ of the found boundary, how many pixels from the found boundary are linked by the minimal distance to the same pixel $e$ of the reference boundary. This number is noted $n_{ef}$, {\em e.g.} on figure \ref{fig_n_ef} we have $n_{ef}=3$ for three different $f$. We redefine the well-detection boundary measure by:
\begin{equation}
\label{WDCN}
WDC = \frac{\sum_f DC_f/n_{ef}} {\left(\max_f(DC_f/n_{ef}) . \sum_e W_e\right)^a}.
\end{equation}

The problem is that the number $n_{ef}$ does not adequately represent a number of pixels on the same boundary and take into account only orthogonal direction. However this measure gives a good evaluation of the proportion of the found boundaries.

\subsection{False detection boundary measure}
The false detection boundary measure is based on the same principle as the well-detected boundary measure, but the Gaussian-kind distribution of weights must be inversed. Hence we can defined a false detection criterion by:
\begin{equation}
\label{FDC}
FDC_f = 1-DC_f / W_e, 
\end{equation}
where the pixels $f$ and $e$ are linked by the minimal distance $d_{fe}$. As a consequence, the false detection boundary measure can be defined by the normalized false detection criterion by:
\begin{eqnarray}
\label{FD}
FD = 1- \exp \left(-\frac{\sum_f\left(FDC_f . n_{ef}\right)}{\max_f(FDC_f . n_{ef}) . \sum_e W_e}\right).
\end{eqnarray}

Here we have described the two measures $FD$ and $WDC$ that compare two images: one image classified by the algorithm and the other one provided by only one expert. In order to evaluate image segmentation algorithms on many images and/or fuse the expert opinions, we can use a weighted sum of these both measures. The weights are given by the image sizes, which can be different for all considered images.

\section{Fusion of classifiers of sonar images}
\label{illustration}
We present here our image classification and segmentation evaluation in a fusion of classifiers of sonar images presented in \cite{Martin05}. Indeed, underwater environment is a very uncertain environment and it is particularly important to classify seabed for numerous applications such as Autonomous Underwater Vehicle navigation. In recent sonar works ({\it e.g.} \cite{LeChenadec05,Lianantonakis05}), the classification evaluation is made only by visual comparison of one original image and the classified image. That is not satisfying in order to correctly evaluate image classification and segmentation.

\subsection{Database}
Our database contains 42 sonar images provided by the GESMA (Groupe d'Etudes Sous-Marines de l'Atlantique). These images were obtained with a Klein 5400 lateral sonar with a resolution of 20 to 30 cm in azimuth and 3 cm in range. The sea-bottom depth was between 15 m and 40 m.

\begin{figure}[htb]
\begin{center}
\includegraphics[height=5cm]{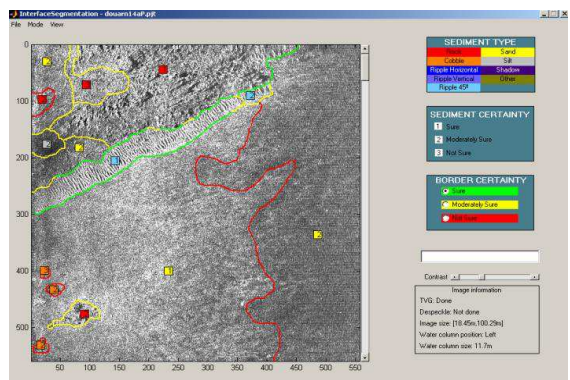}
\end{center}
\caption{Manual Segmentation Interface.}
\label{manual_seg}
\end{figure}

Three experts have manually segmented these images giving the kind of sediment (rock, cobble, sand, silt, ripple (horizontal, vertical or at 45 degrees)), shadow or other (typically ships) parts on images, helped by the manual segmentation interface presented in figure \ref{manual_seg}. All sediments are given with a certainty level (sure, moderately sure or not sure), and the boundary between two sediments is also given with a certainty (sure, moderately sure or not sure). Hence, every pixel of every image is labeled as being either a certain type of sediment or a shadow or other, or a boundary with one of the three certainty levels. We choose the weights: 2/3, 1/2 and 1/3, for respectively the certainty levels: sure, moderately sure and not sure. The proportion of each sediment given by the three experts are given in the table \ref{proportion_sed}. Note that the proportion of the different sediment are very different and that can be a problem for the classification. The proportions are very similar for the three experts. We see that sand and silt are the most present and the shadow and other are very few represented on these images.

\begin{table}[htb]   
\caption{Proportion of sediment in the database (\%)}\label{proportion_sed}
\begin{center}
\begin{tabular}{|c|c|c|c|} \hline
  &     Expert 1  &    Expert 2  & Expert 3 \\ \hline
Rock   & 9.64  & 9.62  & 12.78 \\ \hline
Cobble     &     6.00  & 3.71 & 8.42 \\ \hline
Ripple   &    13.96    &    15.98  & 13.53 \\ \hline
Sand     &     26.97  &    35.62  & 28.40 \\ \hline
Silt     &     42.85  &    34.57  & 35.20 \\ \hline
Shadow     & 0.55  &    0.44  & 0.26 \\ \hline
Other     &    0.10   &    0.05   & 1.40 \\ \hline
\end{tabular}
\end{center}
\end{table}

\subsection{Fusion approaches}

We consider here four methods of features extraction based on four representations of the image: co-occurrence matrices, run-lengths matrix, wavelet transform and Gabor filters \cite{Martin05}. They provide respectively 24, 20, 63 and 4 parameters. These four feature sets are independently considered as the inputs of a multilayer perceptron (MLP) classifier presented in \cite{Martin05}. In order to illustrate our evaluation approach only two of the classifiers fusion presented in \cite{Martin05} are considered coming from the evidence theory. 

The evidence theory is based on basic belief assignments (bba) defined by mapping of each subset of the space of discernment $\Theta=\{C_1, Ö, C_n\}$ onto $[0,1]$, such that: 

\begin{equation}
\label{normDST}
\sum_{X\in 2^\Theta} m(X)=1,
\end{equation}                                 
where $m(.)$ represents the bba. 

The principal difficulty is the choice of a bba according to the application. We can consider two types of approaches: one based on a probabilistic model \cite{Appriou01} and another one based on distance transformation \cite{Denoeux95}. Appriou in \cite{Appriou01} proposes two equivalent models based on three axioms. The first one that we use in this article in order to fuse the decisions of the four classifiers is given by:
\begin{eqnarray}
\left\{
\begin{array}{l}
m_{ji} (\{C_i\})(x)= \frac{\alpha_{ij}R_j p(q_j | C_i)}{1+R_j p(q_j | C_i)}\\
\\
m_{ji} (\{C_i\}^c)(x)=\frac{\alpha_{ij}}{1+R_j p(q_j | C_i)}\\
\\
m_{ji}(\Theta)(x)=1-\alpha_{ij} \\
\end{array}
\right.
\end{eqnarray}
where $q_j$ is the $j^{\mbox th}$ classifier (supposed cognitively independent), $j=1,...,m$,  $\alpha_{ij}$ are reliability coefficients on each classifier $j$ for each class $i=1,...,n$ (in our application we take  $\alpha_{ij}=1$), and $R_j =\left(\max_{q_j,i}p(q_j | C_i)\right)^{-1}$. 

The approach proposed in \cite{Denoeux95} is used in order to fuse the numerical outputs of the four classifiers. The bba are given by:
\begin{eqnarray}
\left\{
\begin{array}{l}
m_{ji} (\{C_i\}/x^{(t)})(x)= \alpha_{ij}\varphi_i(d^{(t)})\\
\\
m_{ji}(\Theta/x^{(t)})(x)=1-\alpha_{ij}\varphi_i(d^{(t)}) \\
\end{array}
\right.
\end{eqnarray}
where $\left(x^{(t)}\right)$ is a set of learning vectors, $d^{(t)}=d(x,x^{(t)})$ is a distance between $x$ and $x^{(t)}$ and $C_i$ is the class of $x^{(t)}$. $\varphi_i$ is a distance function given by:
\begin{eqnarray}
\varphi_i(d)=\exp(-\nu_i d^2),
\end{eqnarray}
where $\nu_i$ is a positive parameter associated to the class $C_i$. 

The combination of the bba is based on the orthogonal non-normalized Dempster-Shafer's rule given in \cite{Smets90} for all $X \in 2^\Theta$ by:
\begin{eqnarray}
\label{conjunctive}
m(X)=\sum_{Y_1 \cap ... \cap Y_M = X} \prod_{j=1}^M m_j(Y_j),
\end{eqnarray}
where $Y_j \in 2^\Theta$ is the response of the expert $j$, and $m_j(Y_j)$ the associated belief function.
In order to conduct the decision, we consider the maximum of pignistic probability \cite{Smets90b}.

\subsection{Evaluation}
Here, we consider six different classes given by the table \ref{class}. The images are considered as a succession of tiles of size 32$\times$32 pixels. Hence the 42 images provide 38997 tiles, units for the classification. The proportion of the number of different sediments on a tile is given in the table \ref{prop_number} for each expert. These proportions are very similar for the three experts.
\begin{table}[htb]   
\caption{Repartition of the kind of sediment in classes}\label{class}
\begin{center}
\begin{tabular}{|c|c|}
\hline
class & sediment \\
\hline
class 1 & rock \\
\hline
class 2 & cobble \\
\hline
class 3 & ripple\\
\hline
class 4 &  sand \\
\hline
class 5 & silt \\
\hline
class 6 & shadow and other\\
\hline
\end{tabular}
\end{center}
\end{table}

\begin{table}[htb]   
\caption{Proportion of number of different kind of sediments on the tiles (\%)}\label{prop_number}
\begin{center}
\begin{tabular}{|c|c|c|c|} \hline
  &     Expert 1  &    Expert 2  & Expert 3 \\ \hline
1 sediment  & 77.79  & 79.65  & 79.94 \\ \hline
2 sediments    &     20.70  & 19.30 & 19.33 \\ \hline
3 sediments   &    1.48    &    1.03  & 0.72\\ \hline
4 sediments    &     0.04  &    0  & 0 \\ \hline
5 sediments     &     0  &    0  & 0 \\ \hline
6 sediments    & 0  &    0 & 0 \\ \hline
\end{tabular}
\end{center}
\end{table}

The total conflict between the three experts is 0.2244. This conflict comes essentially from the difference of opinion of the experts and not from the tiles with more than one sediment. Indeed, we have a weak {\it auto-conflict} (conflict coming from the combination of the same expert three times). The values of the auto-conflict for the three experts are: 0.0496, 0.0474, and 0.0414.

The database is divided into three parts. The first part composed of 20 images (with only 12505 tiles) is used for the learning step of the multilayer perceptron. A second part of 10 images (composed of 12650 tiles) serves the learning step of both the fusion approaches. The used information for these learning stages are only considered given by one of the three experts (expert 1). The last 12 images (corresponding to 13841 tiles) are used in order to evaluate the classifier fusion methods, considering the information given by the two other experts. 

\begin{figure}[htb]
\vspace{-0.5cm}
\begin{center}
\includegraphics[height=5cm]{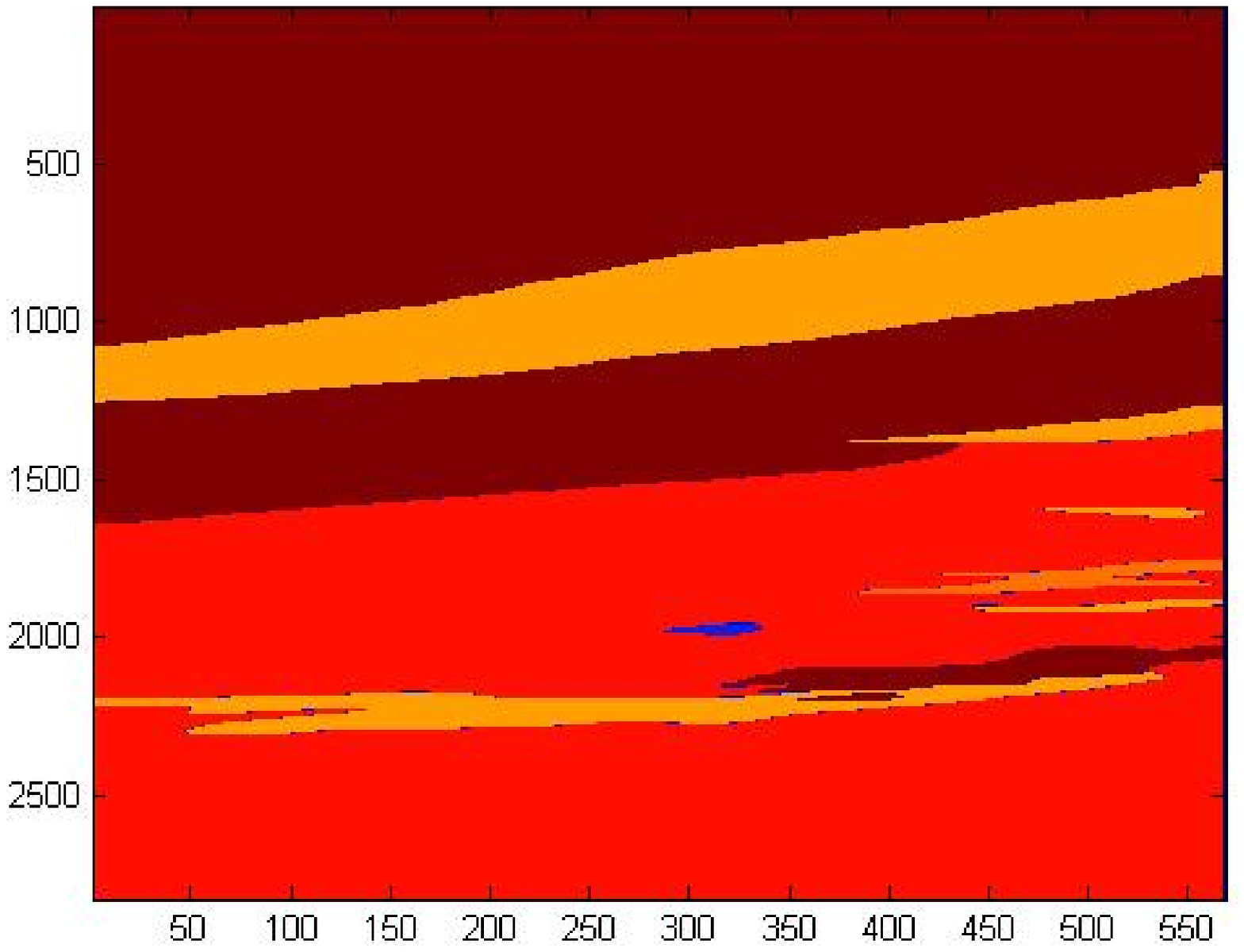}
\includegraphics[height=5cm]{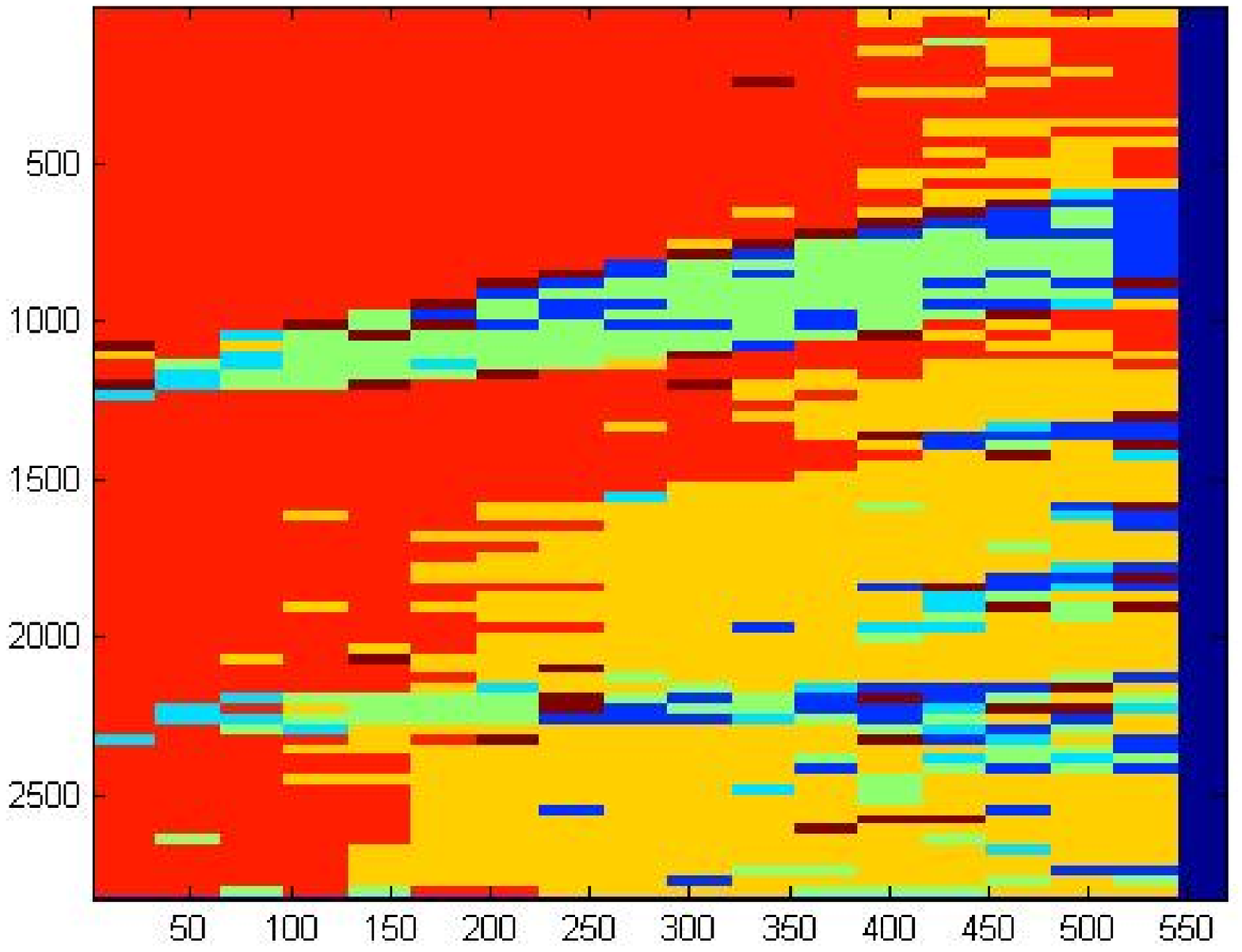}
\includegraphics[height=5cm]{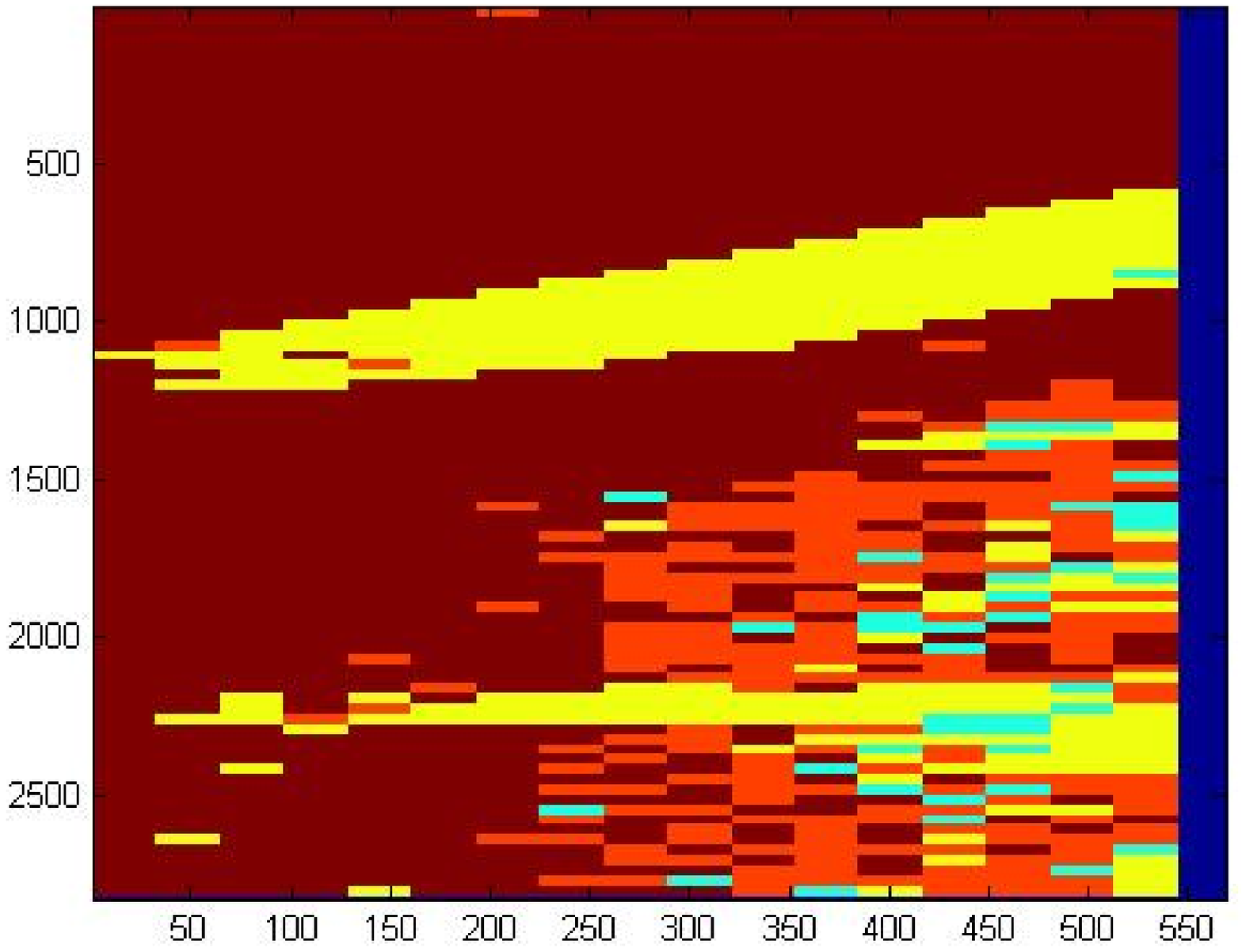}
\end{center}
\vspace{-0.5cm}
\caption{Manual segmentation (first) and automatic segmentation given by the probabilistic approach (second) and the distance approach (third).}
\vspace{-0.5cm}
\label{autoSegIm}
\end{figure}

The figure \ref{autoSegIm} describes the manual segmentation made by one expert and the automatic classification reached by both classifier fusion methods. The dark blue part corresponds to the non considered part of image. First at all if we look on figure \ref{autoSegIm} the results of the classification of the same image, we note that the sediments are quite well classified. However, just looking this figure \ref{autoSegIm} we can not say if the classification is good or not, and if one fusion approach is better or not: it remains very subjective. Moreover it could be good for this image and not for others. So we propose to use our measures. 

First we compare the obtained results to the informations given by only the expert 2. The obtained normalized confusion matrix on the test database is given by for the probabilistic approach:
\begin{eqnarray*}
\label{confusionmatrixApp}
\left\{
\begin{array}{cccccc}
\mbox{rock} & \mbox{cobble} & \mbox{ripple} & \mbox{sand} & \mbox{silt} & \mbox{other} \\
0.00  &  0.01  &  0.00  &  0.01  &  0.01 &  99.97 \\
   13.49  & 24.62  &  0.00 &  33.31  &  0.00  & 28.57 \\
    5.37  &  2.92  & 47.70 &  22.33  &  3.46 &  18.22 \\
    8.78  &  3.10  &  6.97  & 59.51  & 21.05 &  0.59 \\
    0.20  &  0.28  &  0.99  & 16.41 &  82.10  &  0.00 \\
   39.45   &      0    &     0    &   0  & 29.57  & 30.97 \\
\end{array}
\right\}
\end{eqnarray*}
and for the distance approach by:
\begin{eqnarray*}
\label{confusionmatrixDen}
\left\{
\begin{array}{cccccc}
\mbox{rock} & \mbox{cobble} & \mbox{ripple} & \mbox{sand} & \mbox{silt} & \mbox{other} \\
0  &  0.01  & 99.97  &  0.01  &  0.02    &    0 \\
         0 &  32.05 &  20.74 &  34.05 &  13.16   &  0 \\
         0  &  2.90  & 51.51  &  9.28 &  36.31   &  0 \\
         0  &  2.24  &  4.08  & 28.93 &  64.74   &  0 \\
         0  &  0.00  &  0.14  &  4.42 &  95.44   &  0  \\
         0  &      0 &  30.96   &  0 &  69.03    &  0 \\
\end{array}
\right\}
\end{eqnarray*}
We note that the distance approach does not classify rock and other. The most of tiles are classified in ripple and silt and few in sand. The probabilistic approach provides a full confusion matrix. In order to summarize these results, we can give the vector of good-classification rate and the vector of error classification rate given by [0 24.62 47.70 59.51 82.10 30.97] and [94.30 59.13 54.55 82.84 71.18 148.05] for the probabilistic approach and by [0 32.05 51.51 28.94 95.44 0] and [50.00 64.03 144.84 72.88 149.43 50.00] for the distance approach. We recall that is not a percentage because of the weights. The vector of good-classification rates can provide a mean of good-classification rate. We obtain here 62.43 for the probabilistic approach and 50.55 for the distance approach. These results tend to prove that the probabilistic approach gives better results than the distance approach. We can also study the difference on homogeneous tiles and inhomogeneous tiles. For instance, for the probabilistic-based approach, the normalized confusion matrix on homogeneous tiles is given by:
\begin{eqnarray*}
\label{confusionmatrixAppH}
\left\{
\begin{array}{cccccc}
\mbox{rock} & \mbox{cobble} & \mbox{ripple} & \mbox{sand} & \mbox{silt} & \mbox{other} \\
0    &     0   &      0    &     0      &   0 & 100.00\\
   13.49 &  24.62 &    0 & 33.31    &     0 &  28.58\\
    5.37  & 2.92 &  47.70  & 22.33   & 3.46 &  18.22\\
    8.78  &  3.10  &  6.97 &  59.51 &  21.05 &   0.59\\
    0.21  &  0.28  &  0.99  & 16.41 & 82.10  & 0\\
         0  &  0 &  0 &  0 &  0  &  0\\
         \end{array}
         \right\}
\end{eqnarray*}
and on inhomogeneous tiles:
\begin{eqnarray*}
\label{confusionmatrixAppNoH}
\left\{
\begin{array}{cccccc}
\mbox{rock} & \mbox{cobble} & \mbox{ripple} & \mbox{sand} & \mbox{silt} & \mbox{other} \\
25.50 & 11.12 & 7.41 & 15.09 & 14.14 & 26.73 \\
   20.97 & 17.84 & 8.32 & 34.02 & 8.15 & 10.71 \\
   13.50 & 5.71& 30.29 & 29.73 & 7.75 & 13.02 \\
   11.21 & 5.79 & 11.86 & 44.95 & 21.33 & 4.85 \\
   13.79 &  3.24 & 10.24 &  16.40 & 53.55 &  2.77 \\
   39.46  &  0   &  0 & 0 & 29.58 &  30.97 \\
\end{array}
\right\}
\end{eqnarray*}
We observe an important difference. The good-classification rate is better on the homogeneous tiles (62.43) than on the inhomogeneous tiles (39.99). Hence the classification of the inhomogeneous tiles is a real difficulty.

The figure \ref{autoSegIm} seems to show that the segmentation of the distance approach is better than the probabilistic approach. We have to evaluate the segmentation producted by the classification with our measures. Note that this evaluation is highly depending on the size of the tile, here: 32$\times$32 pixels. Our proposed measures, given respectively by the equations (\ref{WDCN}) and (\ref{FD}) expressed in percentage, provide in the case of probabilistic approach 59.84 for the well-detection criterion and 45.64 for the false alarm criterion, and for the distance approach 57.22 for the well-detection criterion and 48.54 for the false alarm criterion. The well-detection criterion and the false alarm criterion of the probabilistic-based fusion are better than the well-detection criterion of the distance-based fusion. However, we have to take care of both measures that are studying together. Indeed, on the figure \ref{autoSegIm}, the probabilistic-based method provides a lot of boundaries, and so the chance to contain well-detection criterion increases, but the false alarm increases also.

In order to confirm these results, we can fuse easily these measures with the resulted measures obtained with the expert 3. The good-classification rate and error classification rate vectors are respectively given by [26.73 14.54 39.83 60.56 81.83 0] and [61.71 57.01 58.47 109.05 111.89 74.16] for the probabilistic-based method and [0 17.94 48.02 30.09 95.83 0] and [50.00 63.31 70.10 87.40 169.54 50.00] for the distance-based method. The mean of the good-classification rate is 58.39 for the probabilistic-based method and 49.24 for the distance-based method. The results of the segmentation evaluation are given by the well-detection criterion and the false alarm criterion: respectively 62.76 and 54.57 for the probabilistic-based approach and 60.83 and 55.90 for the distance-based approach. The fusion of measures originally from the experts shows that the probabilistic-based method is better than the distance-based method. However the difference is lower than with only one expert.

\section{Conclusions}

We have proposed a new evaluation of the image classification and segmentation based on new measures in uncertain environments. In order to achieve a good evaluation of the image classification, we have seen that a linked study of the classification and of the produced segmentation is necessary. The proposed classification evaluation can be used independently for every kind of uncertain units classification, {\em e.g.} is a basic belief assignment is associated to the units. The proposed segmentation evaluation can be used for all image segmentation approaches and not only for a segmentation produced by a classifier. The proposed confusion matrix takes into account the uncertainty of the expert and also the inhomogeneous units ({\it e.g.} patch-worked images in the case of image classification). Moreover we have defined good-classification and errors classification rates from our confusion matrix. The proposed segmentation evaluation considers good and false detection boundary measures where the subjectivity of the expert is considered by the given uncertainty.

In our proposed evaluation approach, the fusion of experts opinions is made by the fusion of our different measures calculated for each expert. This fusion is made by using a simple sum: the uncertainty is considered directly in our measures. It can be interesting to fuse the informations provided by experts before the evaluation in order to obtain an uncertain and imprecise reality. This new reality can used for instance for learning and also for the evaluation of classifiers.



\end{document}